\documentclass[11pt]{article}
\usepackage{acl2012}
\usepackage{setspace}
\usepackage{times}
\usepackage{latexsym}
\usepackage{amsmath}
\usepackage{graphicx}
\usepackage{multirow}
\usepackage{url}

\usepackage[small,compact]{titlesec}
\usepackage{color}
\usepackage[usenames,dvipsnames]{xcolor}
\usepackage{threeparttable}
\newenvironment{enum}{
\begin{enumerate}
  \footnotesize
    \setlength{\itemsep}{1pt}
    \setlength{\parskip}{0pt}
    \setlength{\parsep}{0pt}
    \setlength{\itemindent}{0cm}

}{\end{enumerate}}

\title{Focused Meeting Summarization via Unsupervised Relation Extraction}

\author{Lu Wang \\
  Department of Computer Science \\
  Cornell University \\
  Ithaca, NY 14853 \\
  {\tt luwang@cs.cornell.edu} \\\And
  Claire Cardie \\
  Department of Computer Science \\
  Cornell University \\
  Ithaca, NY 14853 \\
  {\tt cardie@cs.cornell.edu} \\}

\begin{document}

\maketitle

\begin{abstract}
We present a novel unsupervised framework for focused meeting summarization that views the problem as an instance of relation extraction. We adapt an existing in-domain relation learner \cite{ChenBarzilay} by exploiting a set of task-specific constraints and features. We evaluate the approach on a decision summarization task and show that it outperforms unsupervised utterance-level extractive summarization baselines as well as an existing generic relation-extraction-based summarization method. 
Moreover, our approach produces summaries competitive with those generated by supervised methods in terms of the standard ROUGE score.
\end{abstract}

\section{Introduction}
\label{intro}
For better or worse, meetings play an integral role in most of our
daily lives --- they let us share information and collaborate with
others to solve a problem, to generate ideas, and to weigh options.
Not surprisingly then, there is growing interest in developing
automatic methods for meeting summarization (e.g.,
\newcite{Zechner:2002:ASO:638178.638181}, \newcite{Maskey2005},
\newcite{Galley:2006:SCR:1610075.1610126},
\newcite{Lin:2010:RMF:1858681.1858690},
\newcite{Murray:2010:ITA:1857999.1858131}).
This paper tackles the task of {\em focused meeting summarization} ,
i.e., generating summaries of a particular aspect of a meeting rather
than of the meeting as a whole \cite{MurrayEtAl:book:2011}.  For
example, one might want a summary of just the {\sc decisions} made
during the meeting, the {\sc action items} that emerged, the {\sc
  ideas} discussed, or the {\sc hypotheses} put forth, etc.
Consider, for example, the task of summarizing the decisions
in the dialogue snippet in
Figure~\ref{fig:event_stat}. The figure shows only the {\it
decision-related dialogue acts (DRDAs)} --- utterances associated with
one or more decisions.\footnote{These are similar, but not completely
equivalent, to the {\it decision dialogue acts (DDAs)}
of~\cite{Bui:2009:EDM:1708376.1708410},~\cite{Fernandez},~\cite{Frampton:2009:RDD:1699648.1699659}.}
Each DRDA is labeled numerically according to the decision it
supports; so the first two utterances support {\sc Decision 1} as do
the final two utterances in the snippet.  
Manually constructed {\it decision abstracts} for each decision are
shown at the bottom of the
figure.\footnote{\newcite{murray:generating} show that users much
  prefer {\em abstractive summaries} over extracts when the text to be
  summarized is a conversation.  In particular, extractive summaries
  drawn from group conversations can be confusing to the reader
  without additional context; and the noisy, error-prone, disfluent
  text of speech transcripts is likely to result in extractive
  summaries with low readability.}  These constitute the {\it
  decision-focused summary} for the snippet.

\begin{spacing}{0.9}
\begin{figure}[tbhp]
    \hspace{-0.2cm}
    {\scriptsize
    \setlength{\baselineskip}{0pt}
    \begin{tabular}{|l|}
        \hline
  C: Say the \textcolor{red}{standby button} is quite kinda separate from all the\\ other functions. (1)  \\
  C: \underline{Maybe that could} \textit{be} [{a little \textcolor{red}{apple}}]. (1) \\
  C: \underline{It seems like you're gonna} \textit{have} [{\textcolor{blue}{rubber cases}}], as well as~\\               \textcolor{black}{[}\textcolor{blue}{buttons}\textcolor{black}{]}. (2)\\
  A: [{\textcolor{blue}{Rubber buttons}}] \textit{require} [{\textcolor{blue}{rubber case}}]. (2)\\
  A: \underline{You could} \textit{have} [{your \textcolor{Green}{company} badge}] and [{\textcolor{Green}{logo}}]. (3)\\
  A: \underline{I mean} a lot of um computers for instance like like on the one\\ you've got there, it actually     \textit{has} a sort of um [\textcolor{Green}{stick} on badge]. (3)\\
  C: \underline{Shall we} \textit{go} [{for \textcolor{blue}{single curve}}], just to compromise? (2)\\
  B: \underline{We'll} \textit{go} [{for \textcolor{blue}{single curve}}], yeah. (2)\\
  C: And the \textcolor{blue}{rubber} push \textcolor{blue}{buttons}, \textcolor{blue}{rubber case}. (2) \\
  D: \underline{And then are we going for} sort of [{one \textcolor{red}{button}}] \textcolor{red}{\textit{shaped}} \\     \textcolor{black}{[like a fruit]}. $<$vocalsound$>$ Or veg. (1) \\
  D: Could \textit{be} [{a red \textcolor{red}{apple}}], yeah. (1) \\~\\
  
{\bf Decision Abstracts (Summary)} \\
{\sc Decision 1}: The group decided to make the \textcolor{red}{standby button}\\ in the \textcolor{red}{shape} of an \textcolor{red}{apple}. \\
{\sc Decision 2}: The remote will also feature a \textcolor{blue}{rubber case} and\\ \textcolor{blue}{rubber buttons}, and a \textcolor{blue}{single-curved} design. \\
{\sc Decision 3}: The remote will feature the \textcolor{Green}{company logo},\\ possibly in a \textcolor{Green}{sticker} form. \\
        \hline
    \end{tabular}
    }
    \vspace{-0.2cm}
    \caption{\footnotesize Clip from the AMI meeting corpus \cite{Carletta05theami}. A, B, C and D refer to distinct speakers; 
             the numbers in parentheses indicate the associated meeting decision: {\sc decision 1}, {\sc 2} or {\sc 3}. Also shown is the gold-standard (manual) abstract (summary) for each decision. \textcolor{blue}{Colors} indicate overlapping vocabulary between utterances and the summary. \underline{Underlining}, {\it italics}, and [bracketing] are decscribed in the running text.}
    \label{fig:event_stat}
\end{figure}
\end{spacing}

Notice that many 
portions of the DRDAs are not relevant to the decision itself:
%
they often begin with phrases that identify the
utterance within the discourse as potentially introducing a decision
(e.g., ``Maybe that could be'', ``It seems like you're gonna have''),
but do not themselves describe the decision.
We will refer to this portion of a DRDA (underlined in
Figure~\ref{fig:event_stat}) as the {\bf Decision Cue}.

%
Moreover, the decision cue is generally directly followed by the
actual {\bf Decision Content} (e.g., ``be a little apple'', ``have
rubber cases''). Decision Content phrases are denoted in
Figure~\ref{fig:event_stat} via italics and square brackets. Importantly, it is
just the decision content portion of the utterance that should be
considered for incorporation into the focused summary.

%
%

{\it This paper presents an unsupervised framework for focused meeting
summarization that supports the generation of abstractive
summaries.} (Note that we do not currently generate actual abstracts,
but rather aim to identify those {\bf Content} phrases that should comprise the abstract.)
In contrast to existing approaches to focused meeting summarization
(e.g., \newcite{Purver07detectingand}, \newcite{Fernandez},
\newcite{Bui:2009:EDM:1708376.1708410}),
{\em we view the problem as an {\bf information
  extraction} task and hypothesize that existing methods for
domain-specific relation extraction can be modified to identify
salient phrases for use in generating abstractive
summaries}.

Very generally, information extraction methods identify a lexical
``trigger'' or ``indicator'' that evokes a relation of interest and then employ
syntactic information, often in conjunction with semantic constraints,
to find the ``target phrase'' or ``argument constituent'' to be
extracted.  Relation instances, then, are represented by
{\bf indicator-argument} pairs \cite{ChenBarzilay}.

Figure~\ref{fig:event_stat} shows some possible indicator-argument
pairs for identifying the Decision Content phrases in the
dialogue sample.  Content {\bf indicator} words are shown in {\it italics};
the Decision Content target phrases are the {\bf arguments}.
For example, in the fourth DRDA, ``require" is the indicator,
and ``rubber buttons" and ``rubber case" are both
arguments.
Although not shown in Figure~\ref{fig:event_stat}, it is also possible
to identify relations that correspond to the {\bf Decision Cue} phrases.\footnote{Consider, for example, the phrases underlined in the sixth and seventh DRDAs.
``I mean" and ``shall we" are two typical Decision Cue phrases
where ``mean" and ``shall" are possible indicators with ``I"
and ``we" as their arguments, respectively.}

Specifically, we focus on the task of {\it decision summarization}
and, as in previous work in meeting summarization (e.g.,~\newcite{Fernandez},
\newcite{wang-cardie:2011:SummarizationWorkshop}), assume that all
decision-related utterances (DRDAs) have been identified.  We adapt
the unsupervised relation learning approach of \newcite{ChenBarzilay}
to separately identify relations associated with decision cues
vs.\ the decision content within DRDAs by defining a new set of
task-specific constraints and features to take the place of
the domain-specific constraints and features of the original model.
Output of the system is a set of extracted indicator-argument decision 
content relations (see the ``{\sc Our Method}" sample summary of Table 6) that
can be used as the basis of the decision abstract.

We evaluate the approach (using the AMI
corpus~\cite{Carletta05theami}) under two input settings --- in the
{\bf True Clusterings} setting, we assume that the DRDAs for each
meeting have been perfectly grouped according to the decision(s) each
supports; in the {\bf System Clusterings} setting, an automated system
performs the DRDA-decision pairing.
%
The results show that the relation-based summarization approach outperforms two
extractive summarization baselines that select the longest and the
most representative utterance for each decision, respectively. (ROUGE-1
F score of 37.47\% vs.\ 32.61\% and 33.32\% for the baselines given
the True Clusterings of DRDAs.)
Moreover, our approach performs admirably in
comparison to two supervised learning alternatives (scores of 35.61\%
and 40.87\%) that aim to identify the important {\bf tokens} to
include in the decision abstract given the DRDA clusterings. In contrast
to our approach which is transferable to different domains or tasks, these methods would require labeled data for retraining
for each new meeting corpus.

Finally, in order to compare our approach to another {\it relation-based}
summarization technique, we modify the multi-document summarization
system of \newcite{Hachey:2009:MSU:1699510.1699565} to the
single-document meeting scenario.  Here again, our proposed approach
performs better 
(37.47\% vs. 34.69\%).
Experiments under the System Clusterings setting produce the same
overall results, albeit with lower scores for all of the systems and
baselines.








In the remainder of the paper, we review related work in
Section~\ref{rel-work} and give a high-level description of the relation-based
approach to focused summarization
in Section 3.
Sections 4, 5 and 6 present the modifications to the \newcite{ChenBarzilay}
relation extraction model required for its instantiation for the meeting
summarization task.
Sections 7
and 8 provide our experimental setup and results.

\section{Related Work}
\label{rel-work}

Most research on spoken dialogue summarization attempts to generate
summaries for full dialogues~\cite{MurrayEtAl:book:2011}.  
%
%
%
Only recently, however, has the task of focused summarization, and
decision summarization, in particular, been addressed.
%
\newcite{Fernandez} and \newcite{Bui:2009:EDM:1708376.1708410} employ
supervised learning methods to rank phrases or words 
for inclusion in the decision summary.  
In comparison, \newcite{Fernandez} find that the phrase-based approach 
yields better recall than token-based methods, concluding that phrases 
have the potential to support better summaries. 
Input to their system,
however, is narrowed down (manually) from the full set of DRDAs to the subset
that is useful for summarization.
In addition, they evaluate their system w.r.t.\ informative phrases or words that
have been manually annotated within this DRDA subset. 
We are instead interested in comparing our extracted relations to the abstractive summaries.

%
In contrast to our phrase-based approach,
we previously explored a collection
of supervised and unsupervised learning methods for utterance-level
(i.e., dialogue act) and token-level decision summarization \cite{wang-cardie:2011:SummarizationWorkshop}. 
We adopt here the two unsupervised baselines (utterance-level summaries) from
that work for use in
our evaluation.  We further employ their supervised summarization methods as
comparison points for token-level summarization, adding additional features for 
consistency with the other approaches in the evaluation.
%
%
\newcite{Murray:2010:ITA:1857999.1858131} develop an integer linear programming
approach for focused summarization at the utterance-level, selecting sentences that
cover more of the entities mentioned in the meeting as determined through
the use of an external ontology. 
%



The most relevant previous work is~\newcite{Hachey:2009:MSU:1699510.1699565}, 
which uses relational
representations to facilitate sentence-ranking for multi-document
summarization. The method utilizes generic relation extraction to
represent the concepts in the documents as relation instances;
summaries are generated based on a set cover algorithm that selects a
subset of the sentences that best cover the weighted concepts.
Thus, the goal of Hachey's approach is sentence extraction rather than
phrase extraction. 
%
%
Although his relation extraction method, like ours (see Section 4), is probabilistic
and unsupervised (he uses Latent Dirichelt Allocation \cite{Blei:2003:LDA:944919.944937}),
the relations are limited to pairs of named-entities, which is
not appropriate for our decision summarization setting.
Nevertheless, we will adapt his approach for comparison with our relation-based
summarization technique and include it for evaluation.
\section{Focused Summarization as Relation Extraction}
\label{task-as-rel-extraction}

Given the 
DRDAs for each meeting
grouped (not necessarily correctly) according to the decisions they
support, we put each cluster of DRDAs (ordered according to time within the cluster) into one ``decision document".  The goal will be to produce one decision abstract
for each such decision document.  We
obtain constituent and dependency parses using the Stanford
parser~\cite{Klein:2003:AUP:1075096.1075150,stanford_dependencies}. 
With the corpus of constituent-parsed decision documents as the input, we
will use and modify~\newcite{ChenBarzilay}'s system to identify
decision cue relations and decision content relations for each
cluster.\footnote{Other unsupervised
  relation learning methods might also be appropriate (e.g., Open
  IE~\cite{Banko07openinformation}), but they generally model relations between
   pairs of entities 
and group relations only according to lexical similarity.}
(Section~\ref{features} will make clear how the learned decision cue relations 
will be used to identify decision content relations.)
The salient decision content relation instances will be
returned as decision summary components.

Designed for in-domain relation discovery from standard written texts
(e.g., newswire), however, the ~\newcite{ChenBarzilay}
system cannot be applied to our task directly. In our setting,
for example, neither the number of relations nor the
relation types is known in advance. 

In the following sections, we describe the modifications needed for
the spoken meeting genre and decision-focused summarization task.
In particular, \newcite{ChenBarzilay} provide two mechanisms that
allow for this type of tailoring: the {\bf feature set} used to
cluster potential relation instances into groups/types, and a set of
{\bf global constraints} that characterize the general qualities
(e.g., syntactic form, prevalence, discourse behavior) of a good
relation for the task.

\section{Model}
In this section, we describe the Chen et al.~(2011) probabilistic relation learning model used for both {\bf Decision Cue} and {\bf Decision Content} relation extraction. 
The parameter estimation and constraint encoding through posterior inference are presented in Section 5. 

The relation learning model takes as input clusters of DRDAs, sorted according to utterance time and concatenated into one decision document. We assume one decision will be made per document. The goal for the model is to explain how the decision documents are generated from the latent relation variables. The posterior regularization technique (Section 5) biases inference to adhere to the declarative constraints on relation instances. 
In general, instead of extracting relation instances strictly satisfying a set of human-written rules, features and constraints are designed to allow the model to reveal diverse relation types and to ensure that the identified relation instances are coherent and meaningful. For each decision document, we select the relation instance with highest probability for each relation type and concatenate them to form the decision summary.

\begin{figure}
\centering
\includegraphics[width=0.4\textwidth]{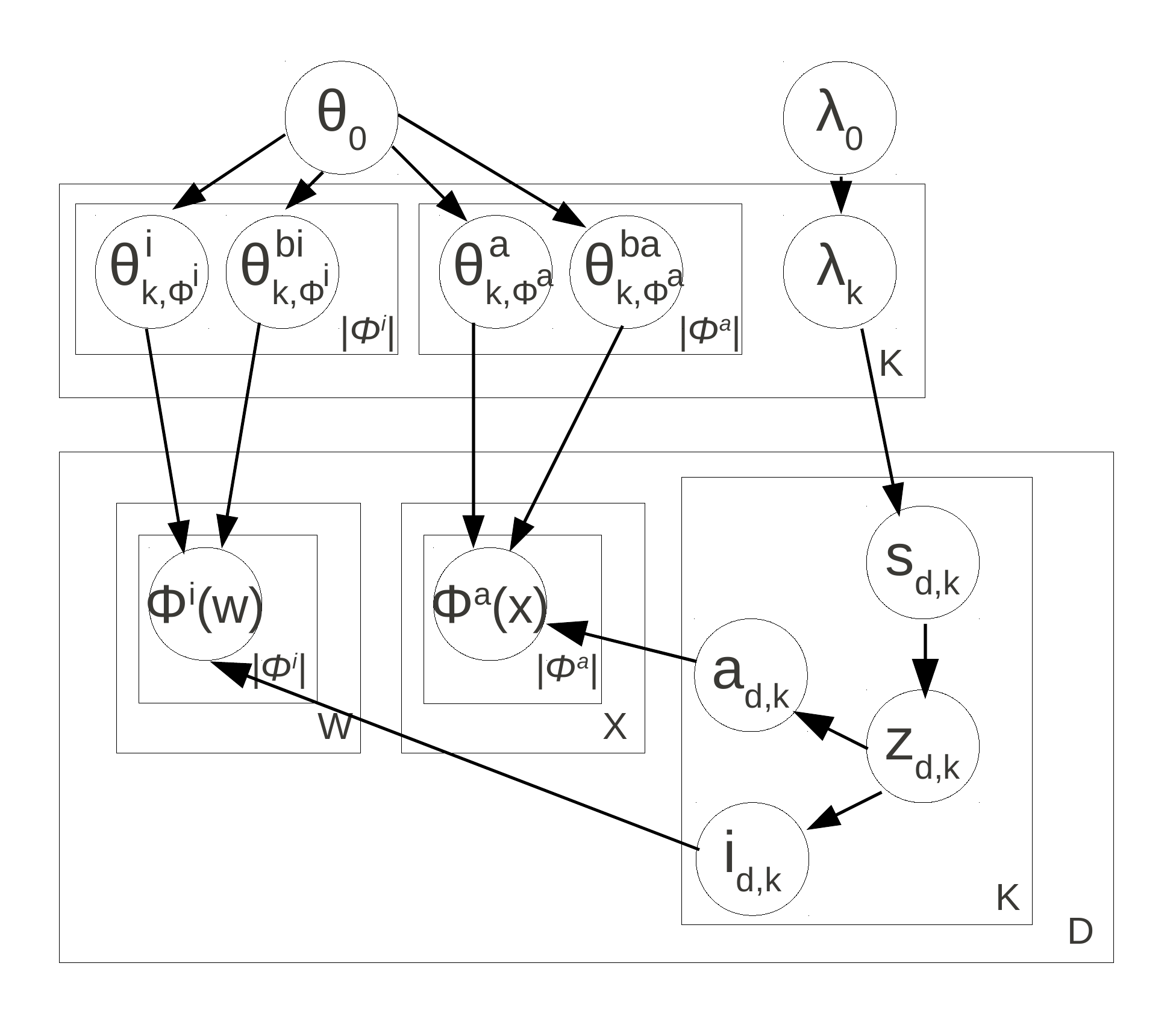}
\vspace{-0.5cm}
\caption{\footnotesize Graphical model representation for the relation learning model. $D$ is the number of decision documents (each decision document consists of a cluster of DRDAs). $K$ is the number of relation types. $W$ and $X$ represent the number of indicators and arguments in the decision document. $|\phi^{i}|$ and $|\phi^{a}|$ are the number of features for indicator and argument.}
\label{fig:plate}
\end{figure}

We restrict the eligible indicators to be a noun or verb, and eligible arguments to be a noun phrase (NP), prepositional phrase (PP) or clause introduced by ``to" (S). 
Given a pre-specified number of relation types $K$, the model employs a set of features $\phi^{i}(w)$ and $\phi^{a}(x)$ (see Section 6) to describe the indicator word $w$ and argument constituent $x$. Each relation type $k$ is associated with a set of {\it feature distributions} $\theta_{k}$ and a {\it location distribution} $\lambda_{k}$. $\theta_{k}$ include four parameter vectors: $\theta_{k}^{i}$ for indicator words, $\theta_{k}^{bi}$ for non-indicator words, $\theta_{k}^{a}$ for argument constituents, and $\theta_{k}^{ba}$ for non-argument constituents. 
Each decision document is divided into $L$ equal-length segments and the location parameter vector $\lambda_{k}$ describes the probability of relation $k$ arising from each segment. The plate diagram for the model is shown in Figure~\ref{fig:plate}. The generative process and likelihood of the model are shown in~\ref{generative}.



\section{Parameter Estimation and Inference via Posterior Regularization}


In order to specify global preferences for the relation instances (e.g.\ the syntactic structure of the expressions), we impose inequality constraints on expectations of the posterior distributions during inference~\cite{NIPS2007_918}. 

\subsection{Variational inference with Constraints}

Suppose we are interested in estimating the posterior distribution $p(\theta, z\vert x)$ of a model in general, where $\theta$, $z$ and $x$ are parameters to estimate, latent variables and observations, respectively. We aim to find a distribution $q(\theta, z)\in Q$ that minimizes the KL-divergence to the true posterior

\vspace{-0.2cm}
{\footnotesize
\begin{equation}
{\textnormal{KL}}(q(\theta, z)\Vert p(\theta, z\vert x))
\end{equation}
}
\vspace{-0.5cm}

A mean-field assumption is made for variational inference, where $q(\theta, z)=q(\theta)q(z)$. Then we can minimize Equation 1 by performing coordinate descent on $q(\theta)$ and $q(z)$.
Now we intend to have fine-level control on the posteriors to induce meaningful semantic parts. For instance, we would like most of the extracted relation instances to satisfy a set of pre-defined syntactic patterns. As presented in~\cite{NIPS2007_918}, a general way to put constraints on posterior $q$ is through bounding expectations of given functions: $E_{q}[f(z)]\leq b$,
where $f(z)$ is a deterministic function of $z$, and $b$ is a pre-specified threshold. For instance, define $f(z)$ as a function to count the number of generated relation instances that meet the pre-defined syntactic patterns, then most of the extracted relation instances will have the desired syntactic structures.

%
%
%
%

By using the mean-field assumption, the model in Section 4 is factorized as

\vspace{-0.3cm}
{\footnotesize
\begin{align}
&q(\theta,\lambda,z,i,a)=\nonumber\\
&\prod_{k=1}^{K}q(\lambda_{k};\hat{\lambda}_{k})q(\theta_{k}^{i};\hat{\theta}_{k}^{i})q(\theta_{k}^{bi};\hat{\theta}_{k}^{bi})q(\theta_{k}^{a}\hat{\theta}_{k}^{a})q(\theta_{k}^{ba};\hat{\theta}_{k}^{ba})\nonumber\\
&\times \prod_{d=1}^{D}q(z_{d,k},i_{d,k},a_{d,k};\hat{c}_{d,k})
\end{align}
}
\vspace{-0.3cm}

The constraints are encoded in the inequalities $E_{q}[f(z,i,a)]\geq b$ or $E_{q}[f(z,i,a)]\leq b$, and affect the inference as described above.  Updates for the parameters are discussed in~\ref{update}.

\subsection{Task-Specific Constraints.}
We define four types of constraints for the decision relation extraction model.
\vspace{-0.1cm}
\paragraph{Syntactic Constraints.}
Syntactic constraints are widely used for information extraction (IE) systems \cite{snow:nips:2005,citeulike:3353876}, as it has been shown that most relations are expressed via a small number of common syntactic patterns. 
For each relation type, we require at least $80\%$\footnote{Experiments show that this threshold is suitable for decision relation extraction, so we adopt it from \cite{ChenBarzilay}.} of the induced relation instances in expectation to match one of the following syntactic patterns:

{\small
\begin{itemize}
\vspace{-0.1cm}
\item The indicator is a verb and the argument is a noun phrase. The headword of the argument is the direct object of the indicator or the nominal subject of the indicator.
\vspace{-0.1cm}
\item The indicator is a verb and the argument is a prepositional phrase or a clause starting with ``to". The indicator and the argument have the same parent in the constituent parsing tree.
\vspace{-0.1cm}
\item The indicator is a noun and is the headword of a noun phrase, and the argument is a prepositional phrase. The noun phrase with the indicator as its headword and the argument have the same parent in the constituent parsing tree.
\end{itemize}
}

For relation $k$, let $f(z_{k},i_{k},a_{k})$ count the number of induced indicator $i_{k}$ and argument $a_{k}$ pairs that match one of the patterns above, and $b$ is set to $0.8D$, where $D$ is the number of decision documents. Then the syntactic constraint is encoded in the inequality $E_{q}[f(z_{k},i_{k},a_{k})]\geq b$.

\vspace{-0.1cm}
\paragraph{Prevalence Constraints.}
The prevalence constraint is enforced on the number of times a relation is instantiated, in order to guarantee that every relation has enough instantiations across the corpus and is task-relevant. Again, we require each relation to have induced instances in at least $80\%$ of decision documents.

\vspace{-0.1cm}
\paragraph{Occurrence Constraints.}
Diversity of relation types is enforced through occurrence constraints. 
In particular, for each decision document, we restrict each word to trigger at most two relation types as indicator and occur at most twice as part of a relation's argument in expectation. An entire span of argument constituent can appear in at most one relation type. 

\vspace{-0.1cm}
\paragraph{Discourse Constraints.}
The discourse constraint captures the insight that the final decision on an issue is generally made, or at least restated, at the end of the decision-related discussion. As each decision document is divided into four equal parts, we restrict $50\%$ of the relation instances to be from the last quarter of the decision documents.



\section{Features}
\label{features}
\begin{spacing}{0.9}
\begin{table}[!t]
\begin{threeparttable}
    {\scriptsize
    \setlength{\baselineskip}{0pt}
    
    \begin{tabular}{|l|}
    \hline
    {\bf Basic Features}\\ \hline
    unigram (stemmed)\\
    part-of-speech (POS)\\
    constituent label (NP, VP, S/SBAR (start with ``to"))\\
    dependency label\\
    \hline
    
  {\bf Meeting Features}\\ \hline
  Dialogue Act (DA) type\\
    speaker role\\
    topic\\    
  \hline    
    
    {\bf Structural Features}~\cite{Galley:2006:SCR:1610075.1610126}~\cite{wang-cardie:2011:SummarizationWorkshop}\\ \hline
    in an Adjacency Pair (AP)?\\ 
    if in an AP, AP type\\
    if in an AP, the other part is decision-related?\\
    if in an AP, the source part or target part?\\
    if in an AP and is source part, is the target positive feedback?\\
    if in an AP and is target part, is the source a question?\\
    \hline
    
    {\bf Semantic Features (from WordNet)}~\cite{Miller:1995:WLD:219717.219748}\\ \hline
  first Synset of head word with the given POS \\
  first hypernym path for the first synset of head word\\
  \hline
    
    {\bf Other Features (only for Argument)}\\ \hline
  number of words (without stopwords)\\
  has capitalized word or not\\
  has proper noun or not\\
    \hline
    
    \end{tabular}
    }
    \vspace{-0.2cm}
    \caption{\footnotesize Features for {\bf Decision Cue} and {\bf Decision Content} relation extraction. All features, except the last type of features, are used for both the indicator and argument. 
\begin{scriptsize}(An Adjacency Pair (AP) is an important conversational analysis concept~\cite{schegloff1973ouc}. In the AMI corpus, an AP pair consists of a source utterance and a target utterance, produced by different speakers.)\end{scriptsize}
    }
    \label{tab:both_feature}
\end{threeparttable}
\end{table}
\end{spacing}

Table~\ref{tab:both_feature} lists the features we use for discovering both the decision cue relations and decision content relations. We start with a collection of domain-independent {\sc Basic Features} shown to be useful in relation extraction~\cite{citeulike:3353876,ChenBarzilay}. 
Then we add {\sc Meeting Features}, {\sc Structural Features} and {\sc Semantic Features} that have been found to be good predictors for decision detection~\cite{Hsueh07whatdecisions} or meeting and decision summarization~\cite{Galley:2006:SCR:1610075.1610126,Murray:2008:SSW:1613715.1613813,Fernandez,wang-cardie:2011:SummarizationWorkshop}. 
Features employed only for argument's are listed in the last category in Table~\ref{tab:both_feature}.

After applying the features in Table~\ref{tab:both_feature} and the global constraints from Section 5 in preliminary experiments, we found that the extracted relation instances are mostly derived from decision cue relations. 
Sample decision cue relations and instances are displayed in Table~\ref{tab:decision_cue_relation_example} and are not necessarily surprising: previous research~\cite{Hsueh07whatdecisions} has observed the 
important role of personal pronouns, such as ``we" and ``I", in decision-making expressions. 
Notably, the decision cue is always followed by the decision content. 
As a result, we include two additional features (see  Table~\ref{tab:decision_content_feature}) that rely on the 
cues to identify the decision content.
Finally,
we disallow content relation instances with an argument containing just a personal pronoun.

\begin{spacing}{0.9}
\begin{table}[!t]
    {\scriptsize
    \setlength{\baselineskip}{0pt}
    \begin{tabular}{|l|l|}
    \hline
    {\bf Decision Cue Relations}&{\bf Relation Instances}\\ \hline
    Group Wrap-up / Recap& we have, we are, we say, we want\\
    Personal Explanation& I mean, I think, I guess, I (would) say\\
    Suggestion& do we, we (could/should) do \\
    Final Decision&it is (gonna), it will, we will\\
    \hline
    
    \end{tabular}
    }
    \vspace{-0.2cm}
    \caption{\footnotesize Sample {\bf Decision Cue} relation instances. The words in parentheses are filled for illustration purposes, while they are not part of the relation instances.}
    \label{tab:decision_cue_relation_example}
\end{table}
\end{spacing}

\begin{spacing}{0.9}
\begin{table}[!t]
    {\scriptsize
    \setlength{\baselineskip}{0pt}
    \begin{tabular}{|l|}
    \hline
    {\bf Discourse Features}\\ \hline
    clause position (first, second, other)\\
  position to the first decision cue relation if any (before, after)\\    
    \hline
    
    \end{tabular}
    }
    \vspace{-0.2cm}
    \caption{\footnotesize Additional features for {\bf Decision Content} relation extraction, inspired by {\bf Decision Cue} relations. Both indicator and argument use those features.}
    \label{tab:decision_content_feature}
\end{table}
\end{spacing}

\section{Experiment Setup}
\label{exp-setup}
\vspace{-0.1cm}
\paragraph{The Corpus.}  We evaluate our approach on the AMI meeting
corpus~\cite{Carletta05theami} that consists of 140 multi-party
meetings with a wide range of annotations. The 129 scenario-driven
meetings involve four participants playing different roles on a design
team. Importantly, the corpus includes a short (usually one-sentence),
manually constructed abstract summarizing each decision discussed in
the meeting.  In addition, all of the dialogue acts that support
(i.e., are relevant to) each decision are annotated as such. We use
the manually constructed decision abstracts as gold-standard
summaries.

\vspace{-0.1cm}
\paragraph{System Inputs.} We consider two system input settings.
In the {\bf True Clusterings} setting, we use the AMI annotations to
create perfect partitionings of the DRDAs for input to the
summarization system; in the {\bf System Clusterings} setting, we
employ a hierarchical agglomerative clustering algorithm used for this
task in previous work \cite{wang-cardie:2011:SummarizationWorkshop}.
The~\newcite{wang-cardie:2011:SummarizationWorkshop} clustering method
groups DRDAs according to their LDA topic distribution similarity.  As
better approaches for DRDA clustering become available, they could be
employed instead.

\vspace{-0.1cm}
\paragraph{Evaluation Metrics.} 
We use the widely accepted ROUGE~\cite{Lin:2003:AES:1073445.1073465} evaluation
measure. 
We adopt the ROUGE-1 and ROUGE-SU4 metrics from~\cite{Hachey:2009:MSU:1699510.1699565}, and also use ROUGE-2.
We choose the stemming option of the ROUGE software at \url{http://berouge.com/} and remove
stopwords from both the system and gold-standard summaries.

\vspace{-0.1cm}
\paragraph{Training and Parameters.}
The Dirichlet hyperparameters are set to 0.1 for the priors. When training the model, ten random restarts are performed and each run stops when reaching a convergence threshold ($10^{-5}$). Then we select the posterior with the lowest final free energy. For the parameters used in posterior constraints, we either adopt them from~\cite{ChenBarzilay} or choose them arbitrarily without tuning in the spirit of making the approach domain-independent.




\vspace{5mm}
We compare our decision summarization approach with (1) two unsupervised
baselines, (2) the unsupervised relation-based approach of \newcite{Hachey:2009:MSU:1699510.1699565},
(3) two supervised methods, and (4) an
upperbound derived from the gold standard decision
abstracts. 

\vspace{-0.1cm}
\paragraph{The {\sc Longest DA} Baseline.} As in~\newcite{Riedhammer:2010:LSS:1837521.1837625} and~\newcite{wang-cardie:2011:SummarizationWorkshop},
this baseline simply selects the longest DRDA in each cluster as the
summary. Thus, this baseline performs utterance-level decision
summarization. Although it's possible that decision content is spread over multiple DRDAs in the cluster, 
this baseline and the next allow
us to determine summary quality when summaries are restricted to a
single utterance.

\vspace{-0.1cm}
\paragraph{The {\sc Prototype DA} Baseline.} Following \newcite{wang-cardie:2011:SummarizationWorkshop},
the second baseline selects the decision cluster prototype (i.e., the
DRDA with the largest TF-IDF similarity with the cluster centroid) as
the summary. 

\vspace{-0.1cm}
\paragraph{The Generic Relation Extraction (GRE) Method of \newcite{Hachey:2009:MSU:1699510.1699565}.} 
\newcite{Hachey:2009:MSU:1699510.1699565} 
presents a generic relation extraction (GRE)
for multi-document summarization. Informative sentences
are extracted to form summaries instead of relation
instances. Relation types are discovered by Latent Dirichlet
Allocation, such that a probability is output for each relation
instance given a topic (equivalent to relation). Their relation
instances are named entity(NE)-mention pairs conforming to a set of
pre-specified rules. For comparison, we use these same rules to select
noun-mention pairs rather than NE-mention pairs, which is
better suited to meetings, which do not contain many NEs.\footnote{Because an approximate set cover algorithm is used in GRE,
one decision-related dialogue act (DRDA) is extracted each time until
the summary reaches the desired length. We run two sets of experiments
using this GRE system with different output summaries --- one selects
one entire DRDA as the final summary
(as~\newcite{Hachey:2009:MSU:1699510.1699565} does), and another one
outputs the relation instances with highest probability conditional on
each relation type. We find that the first set of experiments gets
better performance than the second, so we only report the best results
for their system in this paper.}

\vspace{-0.1cm}
\paragraph{Supervised Learning (SVMs and CRFs).} We also compare our approach 
to two supervised learning methods --- Support
Vector Machines~\cite{citeulike:3340317} with RBF kernel and order-1 Conditional Random
Fields~\cite{Lafferty:2001:CRF:645530.655813} --- trained using the
same features as our system 
(see Tables~\ref{tab:both_feature}
and~\ref{tab:decision_content_feature}) to identify the important {\bf
tokens} to include in the decision abstract. 
Three-fold cross validation is conducted for both methods.

\vspace{-0.1cm}
\paragraph{Upperbound.} We also compute an upperbound that reflects the gap 
between the best possible extractive summaries and the human-written
abstracts according to the ROUGE score: for each cluster of DRDAs, we
select the words that also appear in the associated decision abstract.

\section{Results and Discussion}
\label{results}

\begin{spacing}{0.9}
\begin{table}
    {\scriptsize
    \setlength{\baselineskip}{0pt}
    \begin{tabular}{|c|c|c|c|c|c|}
        \hline
\textbf{ }&\multicolumn{5}{|c|}{\textbf{True Clusterings}}\\
\hline
\textbf{ }&\multicolumn{3}{|c|}{\textbf{R-1}}&\textbf{R-2}&\textbf{R-SU4}\\
\hline
 &PREC&REC&F1&F1&F1\\
    {\bf Baselines}& & & & &\\
  Longest DA& 34.06  &31.28  &32.61&12.03 &13.58\\
  Prototype DA& 40.72  &28.21  &33.32&12.18 &13.46\\
  \hline
    {\bf GRE} & & & & &\\
    5 topics&38.51  &30.66  &34.13& 11.44&13.54\\
  10 topics&39.39  &31.01  &34.69& 11.28&13.42\\
  15 topics&38.00  &29.83  &33.41& 11.40&12.80\\
  20 topics&37.24  &30.13  &33.30& 10.89&12.95\\
  \hline  
    {\bf Supervised Methods} & & & & &\\
    CRF&53.95  &26.57  &35.61 &11.52 &14.07\\
    SVM&42.30  &41.49  & 40.87 &12.91 &16.29\\
  \hline
     {\bf Our Method} & & & & &\\
     5 Relations&39.33  &35.12  &37.10& 12.05&14.29\\
     10 Relations&37.94  &37.03  &{\bf 37.47}& {\bf 12.20}&{\bf 14.59}\\
     15 Relations&37.36  &37.43  &37.39& 11.47&14.00\\
     20 Relations&37.27  &\textbf{\textit{37.64}}  &37.45& 11.40&13.90\\
  \hline  
  {\bf Upperbound} &100.00 & \underline{\textbf{\textit{45.05}}}& 62.12& 33.27&34.89\\  
  
\hline
    \end{tabular}
    }
    \vspace{-0.3cm}
    \caption{\footnotesize ROUGE-1 (R-1), ROUGE-2 (R-2) and ROUGE-SU4 (R-SU4) scores for summaries produced by the baselines, GRE~\cite{Hachey:2009:MSU:1699510.1699565}'s best results, the supervised methods, 
our method and an upperbound --- all with perfect/true DRDA clusterings.}
    \label{tab:rouge comparison results_true}
\end{table}
\end{spacing}

Table~\ref{tab:rouge comparison results_true} illustrates that, using {\bf True (DRDA) Clusterings} our method outperforms the two
baselines and the generic relation extraction (GRE) based system in
terms of F score in ROUGE-1 and ROUGE-SU4 with varied numbers of
relations. Note that for GRE based approach, we only list out their best results for utterance-level summarization. If using the salient relation instances identified by GRE as the summaries, the ROUGE results will be significantly lower.
When measured by ROUGE-2, our method still have better or comparable performances than other unsupervised methods.
Moreover, our system achieves F scores in between those of the 
supervised learning methods, performing better than the CRF in both
 recall and F score. The recall score for the upperbound in ROUGE-1, on the other
hand, indicates that there is still a wide gap between the extractive
summaries and human-written abstracts: without additional lexical information 
(e.g., semantic class information, ontologies) or a real language generation component, 
recall appears to be a bottleneck for
extractive summarization methods that select content only from 
decision-related dialogue acts (DRDAs).

Results using the {\bf System Clusterings} (Table~\ref{tab:rouge comparison results_system}) are comparable, although all
of the system and baseline scores are much lower.
Supervised methods get the best F scores largely due to their high precision;
but our method attains the best recall in ROUGE-1.

\begin{spacing}{0.9}
\begin{table}
    {\scriptsize
    \setlength{\baselineskip}{0pt}
    \begin{tabular}{|c|c|c|c|c|c|}
        \hline
\textbf{ }&\multicolumn{5}{|c|}{\textbf{System Clusterings}}\\
\hline
\textbf{ }&\multicolumn{3}{|c|}{\textbf{R-1}}&\textbf{R-2}&\textbf{R-SU4}\\
\hline
 &PREC&REC&F1&F1&F1\\
    {\bf Baselines}& & & & &\\
  Longest DA&17.06  &11.64  &13.84& 2.76&3.34\\
  Prototype DA&18.14  &10.11  &12.98& 2.84&3.09\\
  \hline
    {\bf GRE} & & & & &\\
    5 topics&17.10  &9.76  &12.40& 3.03&3.41\\
   10 topics&16.28  &10.03  &12.35& 3.00&3.36\\
    15 topics&16.54  &10.90  &13.04& 2.84&3.28\\
    20 topics&17.25  &8.99  &11.80& 2.90&3.23\\
  \hline  
    {\bf Supervised Methods} & & & & &\\
    CRF&47.36  &15.34  &23.18& 6.12&9.21\\
    SVM&39.50  &18.49  & 25.19& 6.15&9.86\\
  \hline
     {\bf Our Method} & & & & &\\
     5 Relations&16.12  &18.93  &17.41& 3.31&5.56\\
  10 Relations&16.27  &18.93  &17.50& 3.32&5.69\\
  15 Relations&16.42  & 19.14  &{\bf 17.68}& {\bf 3.47}&{\bf 5.75}\\
  20 Relations&16.75  &18.25  &17.47& 3.33&5.64\\
  
\hline
    \end{tabular}
    }
    \vspace{-0.3cm}
    \caption{\footnotesize ROUGE-1 (R-1), ROUGE-2 (R-2) and ROUGE-SU4 (R-SU4) scores for summaries produced by the baselines, GRE~\cite{Hachey:2009:MSU:1699510.1699565}'s best results, 
the supervised methods and our method --- all with system clusterings.}
    \label{tab:rouge comparison results_system}
\end{table}
\end{spacing}

%

\paragraph{Discussion.}
To better exemplify the summaries generated by different systems,
sample output for each method is shown in Table~\ref{tab:sample output}. 
The GRE system uses an approximate algorithm for set cover extraction, we list the first three selected DRDA in order.
We see from the table that
utterance-level extractive summaries (Longest DA, Prototype DA,
GRE) make more coherent but still far from concise and compact
abstracts. On the other hand, the supervised methods (SVM, CRF) that
produce token-level extracts better identify the overall content of
the decision abstract.  Unfortunately, they require
human annotation in the training phase; in addition, the output is ungrammatical
and lacks coherence. In comparison, our system presents
the decision summary in the form of phrase-based relations that
provide a relatively comprehensive expression.
\begin{table}[tb]
    \hspace{-0.1cm}
    {\scriptsize
    \setlength{\baselineskip}{0pt}
    \begin{tabular}{|l|}
        \hline

  {\bf DRDA (1)}: Uh the batteries, uh we also thought about that already,\\
  {\bf DRDA (2)}: uh will be chargeable with uh uh an option for a\\ mount station\\
  {\bf DRDA (3)}: Maybe it's better to to include rechargeable batteries\\
  {\bf DRDA (4)}: We already decided that on the previous meeting.\\
  {\bf DRDA (5)}: which you can recharge through the docking station.\\
  {\bf DRDA (6)}: normal plain batteries you can buy at the supermarket\\ or retail shop. Yeah.\\
  \hline
  \hline

  {\bf Decision Abstract}: The remote will use rechargeable batteries\\ which recharge in a docking station.\\
  \hline
  \hline

  {\bf Longest DA \& Prototype DA:}
  normal plain batteries you can \\buy at the supermarket or retail shop. Yeah.\\
  {\bf GRE:}
  1st: normal plain batteries you can buy at the supermarket \\or retail shop. Yeah.\\  
  2nd: which you can recharge through the docking station.\\
  3rd: uh will be chargeable with uh uh an option for a mount station\\
  {\bf SVM:}
  batteries include rechargeable batteries decided recharge\\ docking station\\
  {\bf CRF:}
  chargeable station rechargeable batteries\\
  {\bf Our Method:}
  $<$option, for a mount station$>$,\\ $<$include, rechargeable batteries$>$,\\ $<$decided, that on the previous meeting$>$,\\ $<$recharge, through the docking station$>$,\\ $<$buy, normal plain batteries$>$\\
  \hline
    \end{tabular}
    }
    \vspace{-0.3cm}
    \caption{\footnotesize Sample system outputs by different methods are in the third cell (methods' names are in bold). First cell contains the six DRDAs 
supporting the decision abstracted in the second cell.}
    \label{tab:sample output}
\end{table}

\section{Conclusions}
\label{conclusion}
We present a novel framework for focused meeting summarization based on unsupervised relation extraction. Our approach is shown to outperform unsupervised utterance-level extractive summarization baselines as well as an existing generic relation-extraction-based summarization method.  
Our approach also produces summaries competitive with those generated by supervised methods in terms of the standard ROUGE score. 
Overall, we find that relation-based methods for focused summarization have potential as a technique for supporting the generation of
abstractive decision summaries.


\vspace*{0.5mm}
\begin{small}
\noindent
{\bf Acknowledgments}
This work was supported in part by National Science Foundation Grants
IIS-0968450 and IIS-1111176, and by a gift from Google.
\end{small}

{\scriptsize
\bibliographystyle{acl2012}

}
\appendix
\gdef\thesection{Appendix \Alph{section}}
\section{Generative Process}
\label{generative}
The entire generative process is as follows (``Dir" and ``Mult" refer to the Dirichlet distribution and multinomial distribution):

\begin{enum}
  \item For each relation type $k$:
  \begin{enum}
  \item For each indicator feature $\phi^{i}$, draw feature distributions $\theta_{k,\phi^{i}}^{i}, \theta_{k,\phi^{i}}^{bi}\sim$ Dir$(\theta_{0})$
  \item For each argument feature $\phi^{a}$, draw feature distributions $\theta_{k,\phi^{a}}^{a}, \theta_{k,\phi^{a}}^{ba}\sim$ Dir$(\theta_{0})$
  \item Draw location distribution $\lambda_{k}\sim$ Dir$(\lambda_{0})$
  \end{enum}
  
  \item For each relation type $k$ and decision document $d$:
  \begin{enum}
  \item Select decision document segment $s_{d,k}\sim$ Mult$(\lambda_{k})$
  \item Select DRDA $z_{d,k}$ uniformly from segment $s_{d,k}$, and indicator $i_{d,k}$ and argument constituent $a_{d,k}$ uniformly from DRDA $z_{d,k}$
  \end{enum}
  
  \item For each indicator word $w$ in every decision document $d$:
  \begin{enum}
  \item For each indicator feature $\phi^{i}(w)\sim$ Mult$(\frac{1}{Z}\Pi_{k=1}^{K}\theta_{k,\phi^{i}})$, where $\theta_{k,\phi^{i}}$ is $\theta_{k,\phi^{i}}^{i}$ if $i_{d,k}=w$ and $\theta_{k,\phi^{i}}^{bi}$ otherwise. $Z$ is the normalization factor.
  \end{enum}
  
  \item For each argument constituent $x$ in every decision document $d$:
  \begin{enum}
  \item For each indicator feature $\phi^{a}(x)\sim$ Mult$(\frac{1}{Z}\Pi_{k=1}^{K}\theta_{k,\phi^{a}})$, where $\theta_{k,\phi^{a}}$ is $\theta_{k,\phi^{a}}^{a}$ if $a_{d,k}=x$ and $\theta_{k,\phi^{a}}^{ba}$ otherwise. $Z$ is the normalization factor.
  \end{enum}
\end{enum}

\vspace{-0.3cm}
Given $\theta_{0}$ and $\lambda_{0}$, 
The joint distribution of a set of feature parameters $\boldsymbol{\theta}$, the location distributions $\boldsymbol{\lambda}$, a set of DRDAs $\boldsymbol{z}$, and the selected indicators $\boldsymbol{i}$ and arguments $\boldsymbol{a}$ is:

\vspace{-0.4cm}
{\footnotesize
\begin{align*}
&P(\boldsymbol{\theta, \lambda, z,i,a};\theta_{0},\lambda_{0})=\\
&\prod_{k=1}^{K}P(\theta_{k}^{i};\theta_{0})P(\theta_{k}^{bi};\theta_{0})P(\theta_{k}^{a}|\theta_{0})P(\theta_{k}^{ba};\theta_{0})P(\lambda_{k};\lambda_{0})\\
&\times(\prod_{d=1}^{D}P(i_{d,k};z_{d,k})P(a_{d,k};z_{d,k})P(z_{d,k};s_{d,k})P(s_{d,k};\lambda_{k})\\
&\times (P(w=i_{d,k};\theta_{k}^{i}) \prod_{w\neq i_{d,k}} P(w;\theta_{k}^{bi}))\\
&\times (P(x=a_{d,k};\theta_{k}^{a}) \prod_{x\neq a_{d,k}} P(x;\theta_{k}^{ba})))
\end{align*}
}
%
%

\section{Updates for the Parameters}
\label{update}

The constraints put on the posterior will only affect the update for $q(z)$. For $q(\theta)$, the update is

\vspace{-0.2cm}
{\footnotesize
\begin{equation}
q(\theta)=\arg\!\min_{q(\theta)}{\textnormal{KL}}(q(\theta)\Vert q'(\theta)),
\end{equation}
}
\vspace{-0.3cm}

where $q'(\theta )\propto\exp{\textnormal{E}}_{q(z)}[\log p(\theta,z,x)]$, and $q(\theta)$ is updated to $q'(\theta)$. For $q(z)$, the update is

\vspace{-0.4cm}
{\footnotesize
\begin{align}
&q(z)=\arg\!\min_{q(z)}{\textnormal{KL}}(q(z)\Vert q'(z)) \nonumber\\
&s.t.\ \ E_{q(z)}[f_{c}(z)]\leq b_{c}, \ \forall c\in C
\end{align}
}
\vspace{-0.5cm}

where $q'(z)\propto\exp{\textnormal{E}}_{q(\theta)}[\log p(\theta,z,x)]$. Equation 4 is easily solved via the dual~\cite{NIPS2007_918}~\cite{ChenBarzilay}.

%
%
%
%

\end{document}